\documentclass[10pt,twocolumn,letterpaper]{article}

\usepackage{cvpr}
\usepackage{times}
\usepackage{epsfig}
\usepackage{graphicx}
\usepackage{amsmath}
\usepackage{amssymb}
\usepackage{fancyhdr}

\fancypagestyle{plain}{
	\fancyhf{}
	\fancyhead[C]{IEEE Int. Conf. Computer Vision \& Pattern Recognition (CVPR) as Oral presentation, 2016}    
	\fancyfoot[C]{\thepage}%

}
\usepackage[pagebackref=true,breaklinks=true,letterpaper=true,colorlinks,bookmarks=false]{hyperref}
\usepackage{enumitem}
\usepackage{mathrsfs}
\usepackage{booktabs}

\newcommand{\argmin}{\arg\!\min}

\cvprfinalcopy 

\def\cvprPaperID{165} 

\ifcvprfinal\fi
\begin{document}

\title{Temporally coherent 4D reconstruction of complex dynamic scenes}
\vspace{-0.4cm}
\author{Armin Mustafa  \hspace{.09\linewidth}   Hansung Kim   \hspace{.09\linewidth}   Jean-Yves Guillemaut  \hspace{.09\linewidth}  Adrian Hilton\\
	CVSSP, University of Surrey, Guildford, United Kingdom\\
	{\tt\small a.mustafa@surrey.ac.uk}
}

\maketitle
\vspace{-0.4cm}
\begin{abstract}
\vspace{-0.3cm}
This paper presents an approach for reconstruction of 4D temporally coherent models of complex dynamic scenes.
No prior knowledge is required of scene structure or camera calibration allowing reconstruction from multiple moving cameras.
Sparse-to-dense temporal correspondence is integrated with joint multi-view segmentation and reconstruction  to obtain a complete 4D representation of static and dynamic objects.  
Temporal coherence is exploited to overcome visual ambiguities resulting in improved reconstruction of complex scenes.   
Robust joint segmentation and reconstruction of dynamic objects is achieved by introducing a geodesic star convexity constraint. 
Comparative evaluation is performed on a variety of unstructured indoor and outdoor dynamic scenes with hand-held cameras and multiple people.
This demonstrates reconstruction of complete temporally coherent 4D scene models with improved non-rigid object segmentation and shape reconstruction.
\end{abstract}
\vspace{-0.5cm}
\section{Introduction}
\begin{figure}[t]
	\begin{center}
		\includegraphics[width = 6.3cm] {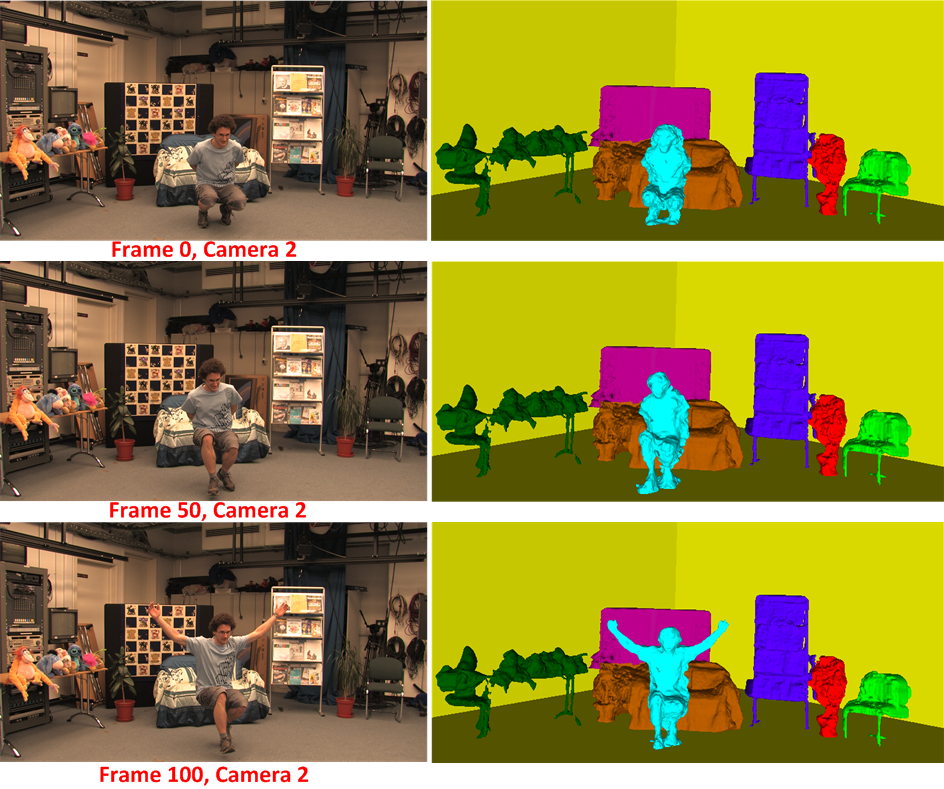}
		\vspace{-0.1cm}
		\caption{Temporally consistent scene reconstruction for Odzemok dataset colour-coded to show the obtained scene segmentation.}
		\vspace{-0.4cm}
		\label{fig:motivation}
	\end{center}
\end{figure}

Existing reconstruction frameworks for general dynamic scenes commonly operate on a frame-by-frame basis \cite{Furukawa10,MustafaICCV15} or are limited to simple scenes \cite{Goldluecke04}.
Previous work on indoor and outdoor dynamic scene reconstruction has shown that joint segmentation and reconstruction across multiple views gives improved reconstruction \cite{Guillemaut2010}.
In this work we build on this concept exploiting temporal coherence of the scene to overcome visual ambiguities inherent in single frame reconstruction and multiple view segmentation methods for general scenes. This is illustrated in Figure \ref{fig:motivation} where the resulting 4D scene reconstruction has temporally coherent labels and surface correspondence for each object. 

We present a sparse-to-dense approach to estimate dense temporal correspondence and surface reconstruction for non-rigid objects.
Initially sparse 3D feature points are robustly tracked from wide-baseline image correspondence using spatio-temporal information to obtain sparse temporal correspondence and reconstruction. 
Sparse 3D feature correspondences are used to constrain optical flow estimation to obtain an initial dense temporally consistent model of dynamic regions. The initial model is then refined using a novel optimisation framework using a geodesic star convexity constraint for simultaneous multi-view segmentation and reconstruction of non-rigid shape.
The proposed approach overcomes limitations of existing methods allowing an unsupervised temporally coherent 4D reconstruction of complete models for general scenes. The scene is automatically decomposed into a set of spatio-temporally coherent objects as shown in Figure \ref*{fig:motivation}.
The contributions are as follows:
\begin{itemize}[topsep=0pt,partopsep=0pt,itemsep=0pt,parsep=0pt] 
	\item Temporally coherent reconstruction of complex dynamic scenes. 
	\item A framework for space-time sparse-to-dense segmentation and reconstruction.
	\item Optimisation of dense reconstruction and segmentation using geodesic star convexity.
	\item Robust and computationally efficient reconstruction of dynamic scenes by exploiting temporal coherence.
\end{itemize}
\vspace{-0.1cm}
\section{Related work}
\label{sec:survey}
\subsection{Temporal multi-view reconstruction}
\vspace{-0.1cm}
Extensive research has been performed in multi-view reconstruction of dynamic scenes. Most existing approaches process each time frame independently due to the difficulty of simultaneously estimating temporal correspondence for non-rigid objects. Independent per-frame reconstruction can result in errors due to the inherent visual ambiguity caused by occlusion and similar object appearance for general scenes. Quantitative evaluation of state-of-the-art techniques for static object reconstruction from multiple views was presented \cite{Seitz06}.
Research investigating spatio-temporal reconstruction across multiple frames \cite{Goldluecke04,Guillemaut3dv} requires accurate initialisation, is limited to simple scenes and does not produce temporally coherent 4D models.
A number of approaches that use temporal information \cite{Bailer15, Cheng09, Larsen07} either require a large number of closely spaced cameras or bi-layer segmentation \cite{Zhang11robustbilayer, Jiang12} as a constraint for complete reconstruction. Other approaches for reconstruction of general scenes from multiple handheld wide-baseline cameras \cite{UnstructuredVBR10, taneja2011modeling} exploit prior reconstruction of the background scene to allow dynamic foreground segmentation and reconstruction. 
Recent approaches for spatio-temporal reconstruction of multi-view data either work on indoor studio data \cite{Oswald14} or for dynamic reconstruction of crowd sourced data \cite{Ji14}.\\
Methods to estimate 3D scene flow have been reported in the literature \cite{Menze2015CVPR}. However existing approaches are limited to narrow baseline correspondence for dynamic scenes. Scene flow approaches dependent on optical flow \cite{Wedel2011, Basha2013} require an accurate estimate for most of the pixels which fails in the case of large motion. 
The approach presented in this paper is for general dynamic indoor or outdoor scenes with large non-rigid motions and no prior knowledge of scene structure. Temporal correspondence and reconstruction are simultaneously estimated to produce a 4D model of the complete scene with both static and dynamic objects.  
%
\vspace{-0.2cm}
\subsection{Multi-view video segmentation}
\vspace{-0.2cm}
In the field of image segmentation, approaches have been proposed to provide impressive temporally consistent video segmentation \cite{GrundmannKwatra2010, Papazoglou13, Narayana13, Zhang13}. Hierarchical segmentation based on graphs was proposed in \cite{GrundmannKwatra2010}, directed acyclic graph were used to propose an object followed by segmentation in \cite{Zhang13} and \cite{Papazoglou13, Narayana13} used optical flow. All of these methods work only for monocular videos.
Recently a number of approaches have been proposed for multi-view foreground object segmentation by exploiting appearance similarity \cite{Djelouah15, Djelouah13, Kowdle12, Lee11, Zeng04} . These approaches assume a static background and different colour distributions for the foreground and background which limits applicability for general complex scenes and non-rigid objects.\\
To address this issue we introduce a novel method for spatio-temporal multi-view segmentation of dynamic scenes using shape constraints. Single image segmentation techniques using shape constraints provide good results for complex scene segmentation \cite{Gulshan10}(convex and concave shapes), but requires manual interaction. The proposed approach performs multi-view video segmentation by initializing the foreground object model using spatio-temporal information from wide-baseline feature correspondence followed by a multi-layer optimization framework using geodesic star convexity to constrain the segmentation. 
Our multi-view formulation naturally enforces coherent segmentation between views and also resolves ambiguities such as the similarity of background and foreground.
\vspace{-0.5cm}
\subsection{Joint segmentation and reconstruction}
\vspace{-0.1cm}
Joint segmentation and reconstruction methods simultaneously estimate multi-view segmentation or matting with reconstruction and have been shown to given improved performance for complex scenes.
A number of approaches have been introduced for joint optimization. However, these are either limited to static scenes \cite{Zach13joint3d, Hane13} or process each frame independently thereby failing to enforce temporal consistency \cite{Campbell201014,MustafaICCV15,Guillemaut2010}. 
A joint formulation for multi-view video was proposed for sports data and indoor sequences in \cite{Guillemaut2010} and for challenging outdoor scenes in \cite{MustafaICCV15}.
Recent work proposed joint reconstruction and segmentation on monocular video achieving semantic segmentation of static scenes. 
Other joint segmentation and reconstruction approaches that use temporal information based on patch refinement \cite{Shin2013, Ozden07} work only for rigid objects. An approach based on optical flow and graph cuts was shown to work well for non-rigid objects in indoor settings but requires silhouettes and is computationally expensive \cite{Guillemaut3dv}.
Practical application of temporally coherent joint estimation requires approaches that work on non-rigid objects for general scenes in uncontrolled environments. 

The proposed approach overcomes the limitations of previous methods enabling robust wide-baseline spatio-temporal  reconstruction and segmentation of general scenes.
Temporal correspondence is exploited to overcome visual ambiguities giving improved reconstruction together with temporally coherent 4D scene models. 
%
\begin{figure*}[t]
	\begin{center}
		\includegraphics[width = 17cm]{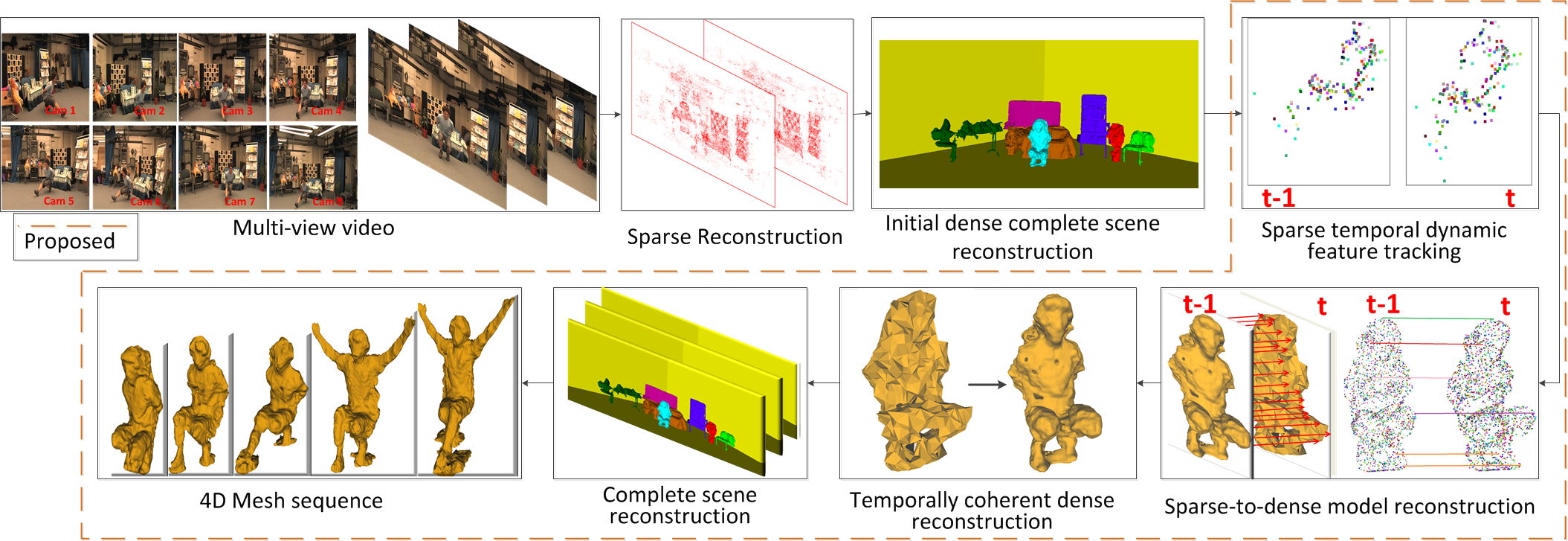}
		\caption{Temporally consistent scene reconstruction framework}
		\vspace{-0.9cm}
		\label{fig:algorithm}
	\end{center}
\end{figure*}
\vspace{-0.2cm}
\section{Methodology}
\label{sec:method}
\vspace{-0.15cm}
This work is motivated by the limitations of existing multiple view reconstruction methods which either work independently at each frame resulting in errors due to visual ambiguity and occlusion \cite{Furukawa10, Guillemaut2010, MustafaICCV15}, or commonly require restrictive assumptions on scene complexity and structure \cite{taneja2011modeling, Guillemaut3dv}.
We address these issues by introducing temporal coherence in the reconstruction to reduce ambiguity, ensure consistent non-rigid structure initialisation at successive frames and improve reconstruction quality.
%
%
\vspace{-0.15cm}
\subsection{Overview}
\label{sec:overview}
\vspace{-0.2cm}
A novel automatic multi-object dynamic segmentation and reconstruction method based on the geodesic star-convexity shape constraint is proposed to obtain a 4D model of the scene including both dynamic and static objects. An overview of the framework is presented in Figure \ref{fig:algorithm} :
\\
\textbf{Sparse reconstruction:}
The input to the system is multiple view wide-baseline video with known camera intrinsics. Extrinsic parameters are calibrated automatically \cite{Hartley03,ImreGH11} using sparse wide-baseline feature matching. 
Segmentation-based feature detection (SFD) \cite{Mustafa15} is used to obtain a relatively large number of sparse features suitable for wide-baseline matching which are distributed throughout the scene including on dynamic objects such as people.
SFD features are matched between views using a SIFT descriptor giving a sparse 3D point-cloud and camera extrinsics  for each time instant. 
The sparse point cloud is clustered in 3D \cite{RusuDoctoralDissertation} with each cluster representing a unique foreground object. Objects with insufficient detected features are reconstructed as part of the scene background. \\
%
%
\textbf{Initial dense complete scene reconstruction:}
Sparse reconstruction at each time instant is clustered in 3D[38] to obtain an initial coarse object segmentation.
Delaunay triangulation \cite{Fortune97} is performed on the set of back projected sparse features for each object in the camera image plane with best visibility. This is propagated to the other views  using the sparse feature matching to obtain an initial object reconstruction.
This reconstruction is refined using the framework explained in Section \ref{sec:optimize} to obtain segmentation and dense reconstruction of each object. \\
Accurate reconstruction of the background object is often challenging due to the lack of features, repetitive texture, occlusion, textureless regions and relatively narrow baseline for distant objects. Hence we create a rough geometric proxy of the background by computing the minimum oriented bounding box for the sparse 3D point cloud using principal component analysis (PCA) \cite{Dimitrov06}. 
The dense reconstruction of the foreground objects and background are combined to obtain a full scene reconstruction at the first time instant. For consecutive time instants only dynamic objects are reconstructed with the segmentation and reconstruction of static objects retained which reduces computational complexity.\\
%
%
\textbf{Temporally coherent reconstruction of dynamic objects:}
Dynamic object regions are detected at each time instant by sparse temporal correspondence of SFD features at successive frames.  
Sparse temporal feature correspondence allows propagation of the dense reconstruction for each dynamic object to obtain an initial approximation  
(Section \ref{sec:InitialReconstruction}). The initial estimate is refined using a joint optimisation of segmentation and reconstruction based on geodesic star convexity (Section \ref{sec:optimize}).
A single 3D model for each dynamic object is obtained by fusion of the view-dependent depth maps using Poisson surface reconstruction \cite{Kazhdan2006}.\\
Subsequent sections present the novel contributions of this work in identifying the dynamic points, initialisation using space-time information and refinement using geodesic star convexity to obtain a dense reconstruction. The approach is demonstrated to outperform state-of-the-art dynamic scene reconstruction and gives a temporally coherent 4D model.
\vspace{-0.2cm}
\subsection{Initial temporally coherent reconstruction} 
\label{sec:InitialReconstruction}
\begin{figure}[t]
	\begin{center}
		\includegraphics[width = 8cm]{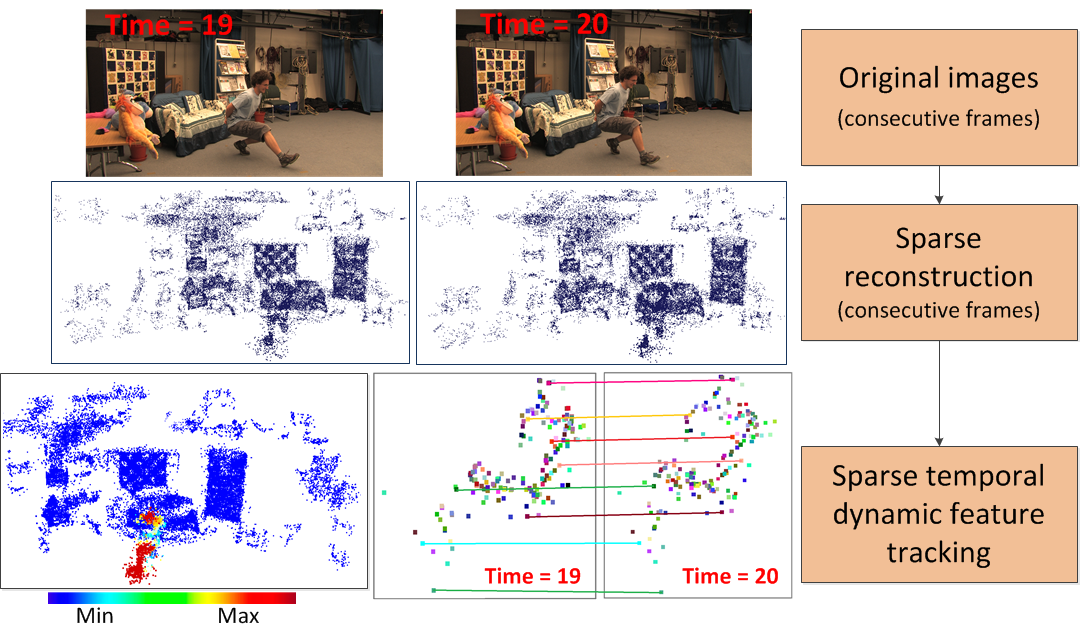}
		\vspace{-0.05cm}
		\caption{Sparse temporal dynamic feature tracking algorithm: Results on two datasets; Min and Max is the minimum and maximum movement in the 3D points respectively.}
		\vspace{-0.3cm}
		\label{fig:dynamicPts}
	\end{center}
\end{figure}

\vspace{-0.2cm}
Once the static scene reconstruction is obtained for the first frame, we perform temporally coherent dynamic scene reconstruction at successive time instants. 
Dynamic regions are identified using temporal correspondence of sparse 3D features. These points are used to obtain an initial dense model for the dynamic objects using optical flow. 
The initial coarse reconstruction for each dynamic region is refined in the subsequent optimization step with respect to each camera view.
Dynamic scene objects are identified from the temporal correspondence of sparse feature points. Sparse correspondence is then used to propagate an initial model of the moving object for refinement. Figure \ref{fig:dynamicPts} presents the sparse reconstruction and temporal correspondence.\\
%
\begin{figure}[t]
	\begin{center}
		\includegraphics[width = 4.7 cm]{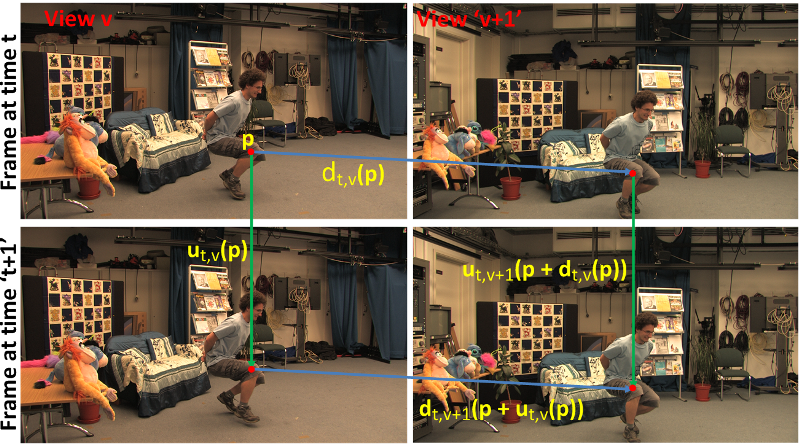}
		\caption{Spatio-temporal consistency check for 3D tracking}
		\vspace{-0.4cm}
		\label{fig:3Dtracking}
	\end{center}
\end{figure}
\begin{figure}[t]
	\begin{center}
		\includegraphics[width = 8 cm]{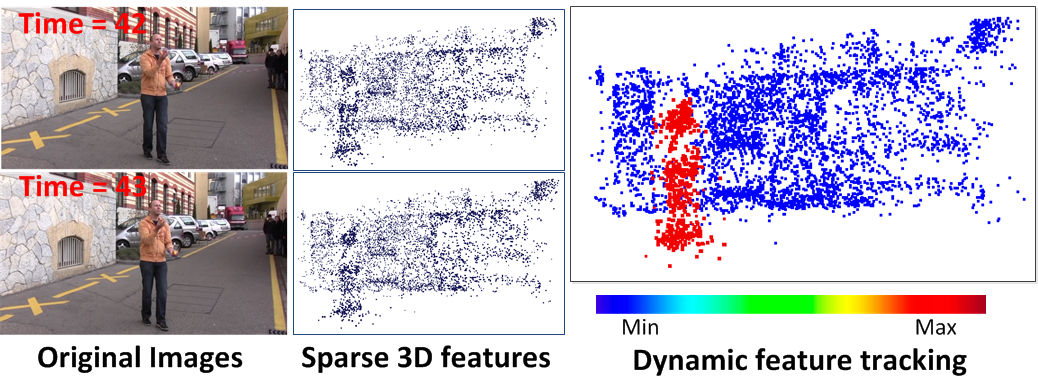}
		\vspace{-0.1cm}
		\caption{Sparse dynamic feature tracking for Juggler dataset.}
		\vspace{-0.5cm}
		\label{fig:3DtrackingJuggler}
	\end{center}
\end{figure}
%
%
\textbf{Sparse temporal dynamic feature tracking:} 
Numerous approaches have been proposed to track moving objects in 2D using either features or optical flow. However these methods may fail in the case of occlusion, movement parallel to the view direction, large motions and moving cameras.
To overcome these limitations we match the sparse 3D feature points obtained using SFD from multiple wide-baseline views at each time instant.
The use of sparse 3D features is robust to large non-rigid motion, occlusions and camera movement. SFD \cite{Mustafa15} detects sparse features which are stable across wide-baseline views and consecutive time instants for a moving camera and dynamic scene.  
Sparse 3D feature matches between consecutive time instants are back-projected to each view.
These features are matched temporally using a SIFT descriptor to identify the moving points.
Robust matching is achieved by enforcing multiple view consistency for the temporal feature correspondence in each view as illustrated in 
Figure \ref{fig:3Dtracking}. 
Each match must satisfy the constraint:
\vspace{-0.3cm}
\begin{eqnarray*}
 \left \| d_{t,v}(p) + u_{t,v+1}(p+d_{t,v}(p)) - u_{t,v}(p) -\right. \\
  \left. d_{t,v+1}(p+u_{t,v}(p)) \right \|< \epsilon  \vspace{-2cm}
 \end{eqnarray*}
where $p$ is the feature image point in view v at frame $t$, $d_{t,v}(p)$ is the disparity at frame $t$ from view $v$ to $v+1$, $u_{t,v}(p)$ is the temporal correspondence from frames $t$ to $t+1$ for view $v$.
The multi-view consistency check ensures that correspondences between any two views remain temporally consistent for  successive frames. 
Matches in the 2D domain are sensitive to camera movement and occlusion, hence we map the set of refined matches into 3D to make the system robust to camera motion. 
The Frobenius norm is applied on the 3D point gradients in all directions \cite{Zhang13} to obtain the `net' motion at each sparse point. The `net' motion between pairs of 3D points for consecutive time instants are ranked, and the top and bottom 5 percentile values removed. Median filtering is then applied to identify the dynamic features. Figure \ref{fig:3DtrackingJuggler} shows an example with moving cameras.
\\
\begin{figure}[b]
	\begin{center}
		\includegraphics[width = 8.5cm]{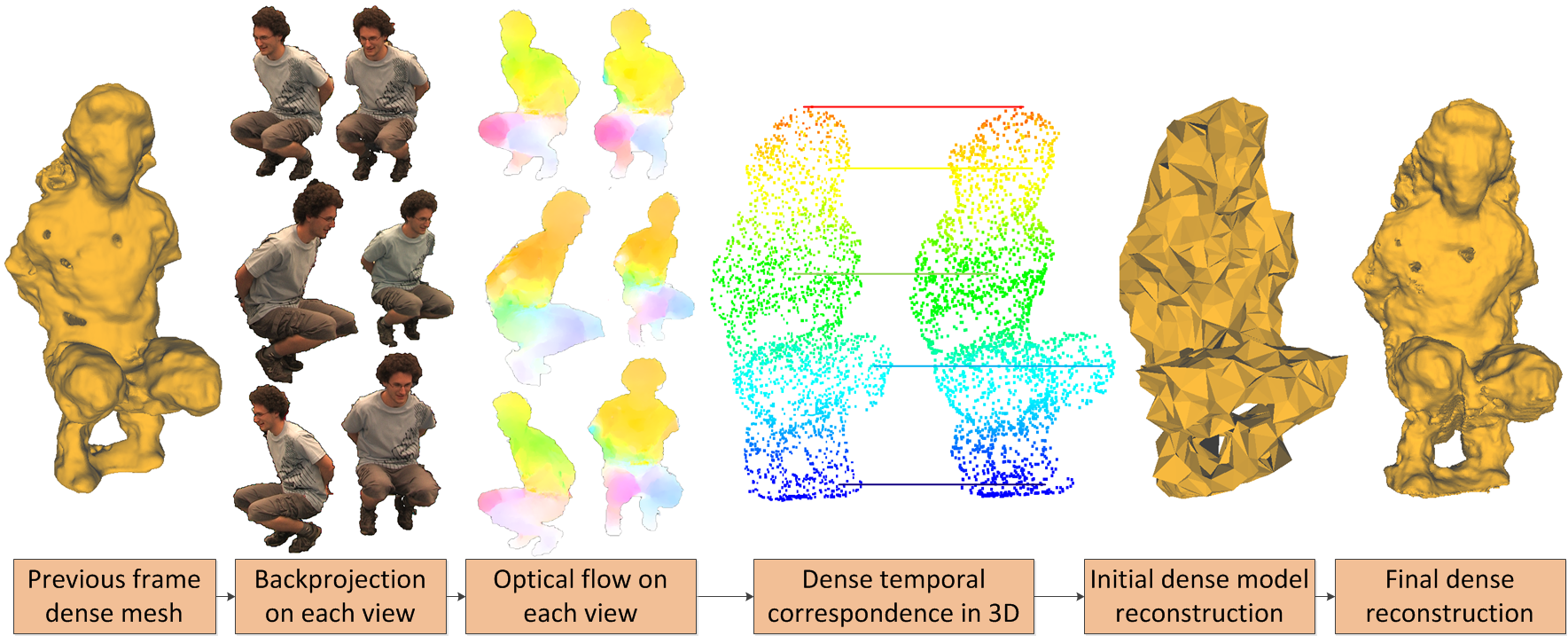}
		\vspace{-0.5cm}
		\caption{Initial sparse-to-dense model reconstruction workflow}
		\vspace{-0.6cm}
		\label{fig:initialmodel}
	\end{center}
\end{figure}
%
\noindent
\textbf{Sparse-to-dense model reconstruction:} 
Dynamic 3D feature points are used to initialize the segmentation and reconstruction of the initial model. 
This avoids the assumption of static backgrounds and prior scene segmentation commonly used to initialise multiple view reconstruction with a coarse visual-hull approximation \cite{Guillemaut2010}. 
Temporal coherence also provides a more accurate initialisation to overcome visual ambiguities at individual frames. 
Figure \ref{fig:initialmodel}  illustrates the use of temporal coherence for reconstruction initialisation and refinement.
Dynamic feature correspondence is used to identify the mesh for each dynamic object. This mesh is back projected on each view to obtain the region of interest. 
Optical flow \cite{Bouguet00} is performed on the projected mask for each view in the temporal domain using the dynamic feature correspondences over time as initialization. 
Dense multi-view wide-baseline correspondences from the previous frame are propagated to the current frame using the information from the flow vectors to obtain dense multi-view matches in the current frame. 
The matches are triangulated in 3D to obtain a refined 3D dense model of the dynamic object for the current frame.\\
For dynamic scenes, a new object may enter the scene or a new part may appear as the object moves. To allow the introduction of new objects and object parts we also use information from the cluster of sparse points for each dynamic object. The cluster corresponding to the dynamic features is identified and static points are removed. This ensures that the set of new points not only contain the dynamic features but also the unprocessed points which represent new parts of the object. These points are added to the refined sparse model of the dynamic object. To handle the new objects we detect new clusters at each time instant and consider them as dynamic regions.\\
Once we have a set of dense 3D points for each dynamic object, Poisson surface reconstruction is performed on the set of sparse points to obtain an initial coarse model of each dynamic region $\mathscr{R}$, which is subsequently refined using the optimization framework (Section \ref{sec:optimize}). 
%
%
\subsection{Temporally coherent dense reconstruction}
\label{sec:optimize}
\begin{figure}[t]
	\begin{center}
		\includegraphics[width = 7cm]{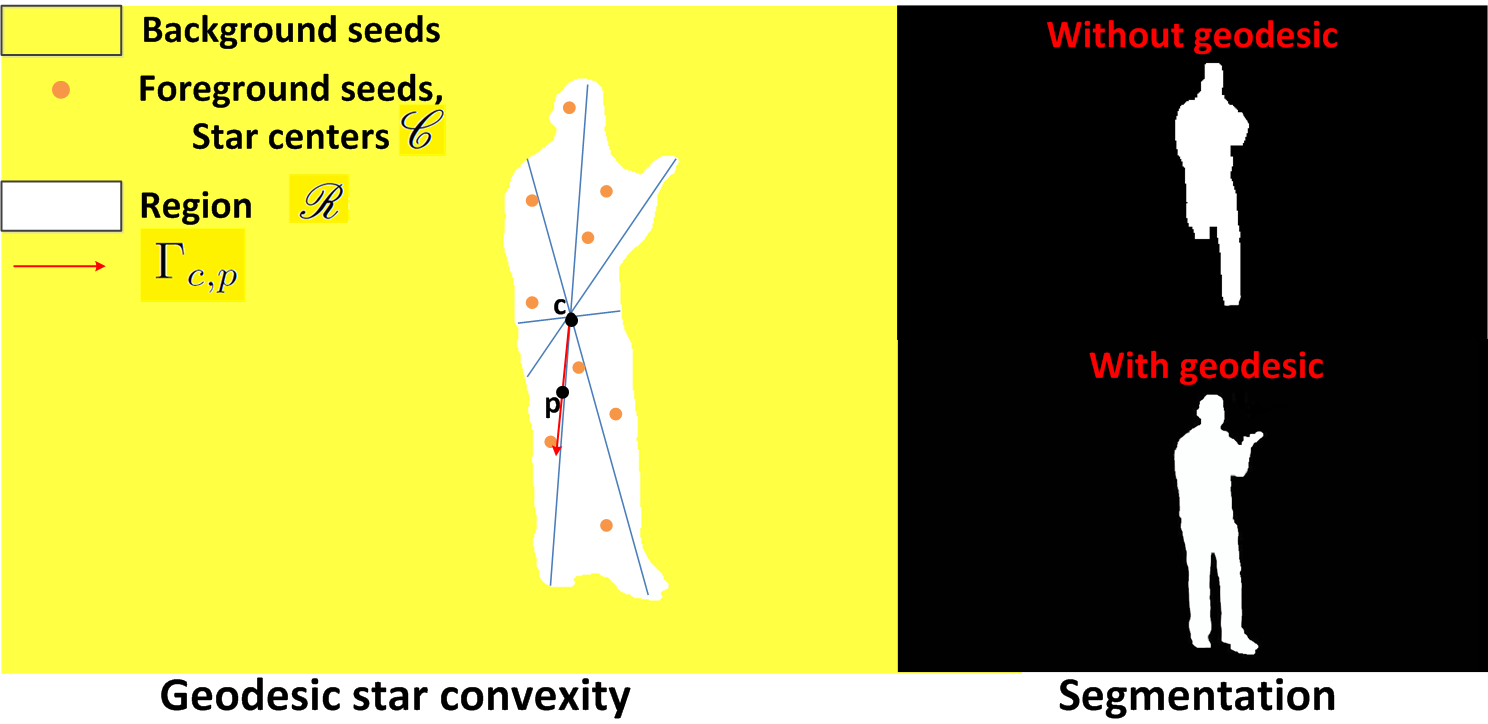}
		\vspace{-0.1cm}
		\caption{Geodesic star convexity: A region $\mathscr{R}$ with star centers $\mathscr{C}$ connected with geodesic distance $\Gamma _{c,p}$. Segmentation results with and without geodesic star convexity based optimization are shown on the right for the Juggler dataset.}
		\vspace{-0.4cm}
		\label{fig:geodesic}
	\end{center}
\end{figure}

The initial reconstruction and segmentation from dense temporal feature correspondence is refined using a joint optimization framework. A novel shape constraint is introduced based on geodesic star convexity which has previously been shown to give improved performance in interactive image segmentation for structures with fine details (for example a persons fingers or hair)\cite{Gulshan10}. In this work the shape constraint is automatically initialised for each view from the initial segmentation.
The geodesic star-convexity is integrated as a constraint on the energy minimisation for joint multi-view reconstruction and segmentation \cite{Guillemaut2010}. The shape constraint is based on the geodesic distance with foreground object initialisation (seeds) as star centres to which the object shape is restricted. The union formed by multiple object seeds form a geodesic forest. This allows complex shapes to be segmented. 
In this work to automatically initialize the segmentation we use the sparse temporal feature correspondence as star centers (seeds) to build a geodesic forest automatically. The region outside the initial coarse reconstruction of all dynamic objects is initialized as the background seed for segmentation as shown in in Figure \ref{fig:geodesic}. The shape of the dynamic object is restricted by this geodesic distance constraint that depends on the image gradient.
Comparison with existing methods for multi-view segmentation demonstrates improvements in recovery of fine detail structure as illustrated in Figure \ref{fig:geodesic}.
%
\vspace{-0.2cm}
\subsubsection{Optimization based on geodesic star convexity}
\label{subsec:problem}
\vspace{-0.1cm}
The depth of the initial coarse reconstruction estimate is refined for each dynamic object at a per pixel level.
Our goal is to assign an accurate depth value from a set of depth values $\mathscr{D} = \left \{ d_{1},...,d_{\left|\mathscr{D} \right|-1} , \mathscr{U} \right \}$ and assign a layer label from a set of label values $\mathscr{L} = \left \{ l_{1},...,l_{\left|\mathscr{L} \right|} \right \}$ to each pixel $p$ for the region $\mathscr{R}$ of each dynamic object.
Each $d_{i}$ is obtained by sampling the optical ray from the camera and $\mathscr{U}$ is an unknown depth value to handle occlusions.
This is achieved by optimisation of a joint cost function \cite{Guillemaut2010} for label (segmentation) and depth (reconstruction): \\
$ E(l,d) = \lambda _{data}E_{data}(d) + \lambda _{contrast}E_{contrast}(l)  +$
\vspace{-0.2cm}
\begin{equation} \label{eq:costfunction}
	\hspace{-2.8cm} \lambda _{smooth}E_{smooth}(l,d) + \lambda _{color}E_{color}(l) 
	\vspace{-0.2cm}
\end{equation}
where, $d$ is the depth at each pixel, $l$ is the layer label for multiple objects and the cost function terms are defined in section \ref{sec:cost}. 
This is solved subject to a geodesic star-convexity constraint on the labels $l$.
\label{sec:dense}
A label $l$ is star convex with respect to center $c_{i}$, if every point $p\in l$ is visible to a star center $c_{i}$ in set $\mathscr{C} = \left \{ c_{1},...,c_{n}  \right \}$ via $l$ in the image $x$, where $n$ is the number of star centers\cite{Gulshan10}. This is expressed as an energy cost:
\vspace{-0.2cm}
\begin{equation} \label{eq:geodesic1}
	E^{\star}(l \rvert x, \mathscr{C}) = \sum_{p\in \mathscr{R}} \sum_{q \in \Gamma _{c,p}} E_{p,q}^{\star}(l_p , l_q)
	\vspace{-0.2cm}
\end{equation}
\begin{equation} \label{eq:geodesic2}
	\forall q \in \Gamma _{c,p} , \text{      } E_{p,q}^{\star} = \left\{\begin{matrix}
		\infty \text{          if  } l_p \neq l_q\\ 
		0      \text{                      otherwise              }\\ 
	\end{matrix}\right.
\end{equation}
where $\forall p \in \mathscr{R}: p \in l \Leftrightarrow l_p = 1 $ and $ \Gamma _{c,p}$ is the geodesic path joining $p$ to any star center in set $\mathscr{C}$ given by:
\vspace{-0.2cm}
\begin{equation} \label{eq:geodesic3}
	\Gamma _{c,p} = \argmin_{\Gamma \in \mathscr{P}_{c,p}} \mathscr{L}(\Gamma)
	\vspace{-0.2cm}
\end{equation}
where $\mathscr{P}_{c,p}$ denotes the set of all discrete paths between $c$ and $p$ and $\mathscr{L}(\Gamma)$ is the length of discrete geodesic path as defined in \cite{Gulshan10}.
In our case we define the temporal sparse feature correspondences as star centers, hence the segmentation will include all the points which are visible to these sparse features via geodesic distances in the region $\mathscr{R}$, thereby employing the shape constraint. Since the star centers are selected automatically, the method is unsupervised.  
The energy in the Eq. \ref{eq:costfunction} is minimized as follows:
\vspace{-0.3cm}
\begin{equation} \label{eq:minimize}
	\underset{s.t.}{min_{(l,d)}} \text{         }\underset{l\epsilon S^{\star }(\mathscr{C})}{E(l,d)}   \Leftrightarrow \min_{(l,d)}E(l,d) + 	E^{\star}(l \rvert x, \mathscr{C})
	\vspace{-0.2cm}
\end{equation}
where $ S^{\star }(\mathscr{C})$ is the set of all shapes which lie within the geodesic distances wrt to the centers in $\mathscr{C}$.
Optimization of eq. \ref{eq:minimize}, subject to each pixel $p$ in the region $\mathscr{R}$ being at a geodesic distance from the star centers in the set $\mathscr{C}$, is performed using the $\alpha$-expansion algorithm for a pixel $p$ by iterating through the set of labels in $\mathscr{L} \times \mathscr{D}$ \cite{Boykov01}. Graph-cut is used to obtain a local optimum \cite{Kolmogorov04}. 
\vspace{-0.3cm}
\subsubsection{Energy cost function}\label{sec:cost}
\vspace{-0.1cm}
For completeness in this section we define each of the terms in eq. \ref{eq:costfunction}, these are based on previous terms used for joint optimisation over depth for each pixel introduced in \cite{MustafaICCV15}, with modification of the color matching term to improve robustness and extension to multiple labels. \\
\textbf{Matching term:}
The data term for matching between views is specified as a measure of photo-consistency as follows:\\
$ E_{data}(d) = \sum_{p\in \mathscr{P}} e_{data}(p, d_{p}) = $
\vspace{-0.3cm}
\begin{equation} \label{eq:matching1}
	\hspace{-0.75cm}
	\begin{cases}
		M(p, q)  = \sum_{i \in \mathscr{O}_{k}}m(p,q) ,& \text{if } d_{p}\neq \mathscr{U}\\
		M_{\mathscr{U}}, & \text{if } d_{p} = \mathscr{U}\\
	\end{cases}
	\vspace{-0.3cm}
\end{equation}
where $\mathscr{P}$ is the 4-connected neighbourhood of pixel $p$, $M_{\mathscr{U}}$ is the fixed cost of labelling a pixel unknown and $q$ denotes the projection of the hypothesised point $P$ in an auxiliary camera where $P$ is a $3D$ point along the optical ray passing through pixel $p$ located at a distance $d_{p}$ from the reference camera. $\mathscr{O}_{k}$ is the set of $k$ most photo-consistent pairs with reference camera and $m(p,q)$ is inspired from \cite{Hu12}.\\
%
\textbf{Contrast term:}
The contrast term is as follows:
\vspace{-0.2cm}
\begin{equation} \label{eq:contrast1}
	E_{contrast}(l) =  \sum_{p,q \in \mathscr{N}} e_{contrast}(p,q,l_p,l_q)
\end{equation}
\vspace{-0.3cm}
\begin{equation} \label{eq:contrast2}
	e_{contrast}(p,q,l_p,l_q)=  
	\begin{cases}
		0, & \text{if } (l_{p} =  l_{q})\\
		\frac{1}{1+\epsilon }( \epsilon + exp^{-C(p,q)}),  & \text{otherwise}
	\end{cases}
\end{equation}
%
\textbf{Smoothness term:}
This term is defined as:
\vspace{-0.2cm}
\begin{equation} \label{eq:smooth}
	E_{smooth}(l,d) = \sum_{(p,q)\in \mathscr{N}} e_{smooth}(l_p,d_{p},l_q,d_{q})
	\vspace{-0.3cm}
\end{equation}
$e_{smooth}(l_p,d_{p},l_q,d_{q})=$
\vspace{-0.3cm}
\begin{equation}
	\begin{cases}
		min(\left | d_{p} - d_{q} \right |, d_{max}),& \text{if } l_{p} =  l_{q} \text{ and }  d_{p},d_{q}\neq \mathscr{U}\\
		0,              & \text{if } l_{p} =  l_{q} \text{ and } d_{p},d_{q} = \mathscr{U}\\
		d_{max},  & \text{otherwise}
			\vspace{-0.4cm}
	\end{cases}
\end{equation}
$d_{max}$ is set to 50 times the size of the depth sampling step defined in Section \ref{subsec:problem} for all datasets.\\
%
%
\textbf{Color term:}
This term is computed using the negative log likelihood \cite{Kolmogorov04} of the color models learned from the foreground and background markers. The star centers obtained from the sparse 3D features are foreground markers and for background markers we consider the region outside the projected initial coarse reconstruction for each view. The color models use GMMs with 5 components each for FG/BG mixed with uniform color models \cite{Das09} as the markers are sparse. 
\vspace{-0.3cm}
\begin{equation} \label{eq:color}
	E_{color}(l) = \sum_{p\in \mathscr{P}} -log P(I_{p}\rvert l_{p})
	\vspace{-0.3cm}
\end{equation}
where $P(I_{p}\rvert l_{p} = l_i)$ denotes the probability at pixel $p$ in the reference image belonging to layer $l_i$.\\
\begin{figure}[t]
	\begin{center}
		\includegraphics[width = 8 cm]{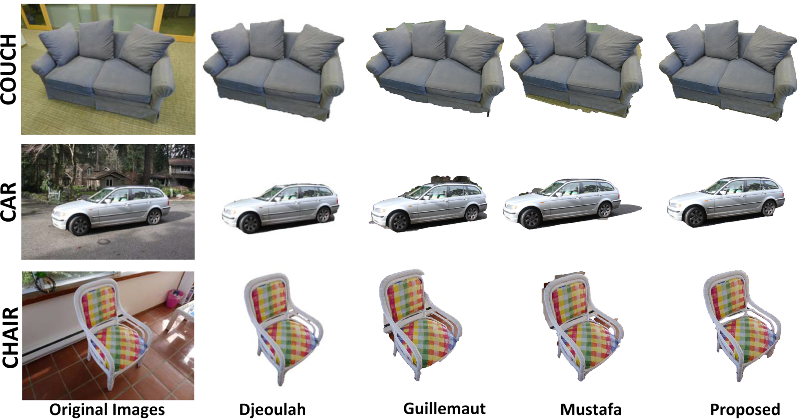}
		\vspace{-0.1cm}
		\caption{Comparison of segmentation on benchmark static datasets using geodesic star-convexity.}
		\vspace{-0.5cm}
		\label{fig:staticSeg}
	\end{center}
\end{figure}
\begin{figure}[t]
	\begin{center}
		\includegraphics[width=0.99\linewidth]{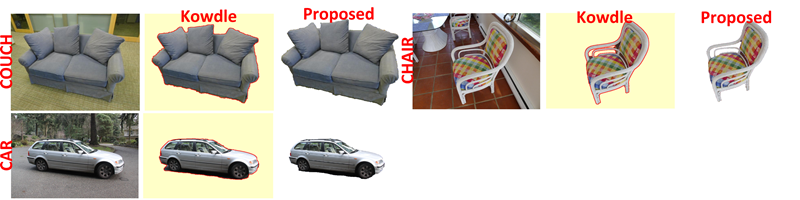}
	\end{center}
	\vspace{-0.5cm}
	\caption{Comparison of segmentation with Kowdle.}
	\label{fig:kowdle}
\end{figure}
\begin{figure}[t]
	\begin{center}
		\includegraphics[width = 8 cm]{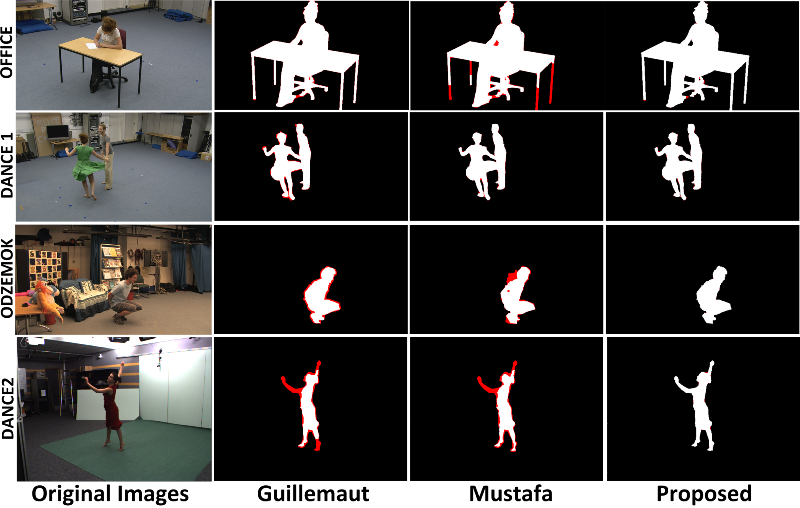}
		\vspace{-0.1cm}
		\caption{Segmentation results for dynamic scenes  (Error against ground-truth is highlighted in red).}
		\vspace{-0.4cm}
		\label{fig:segment}
	\end{center}
\end{figure}
\begin{figure*}[t]
	\begin{center}
		\includegraphics[width = 16cm]{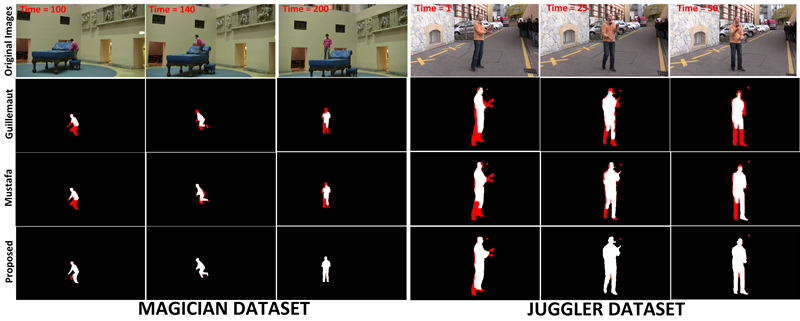}
		\vspace{-0.25cm}
		\caption{Segmentation results for dynamic scenes on sequence of frames (Error against ground-truth is highlighted in red).}
		\vspace{-0.7cm}
		\label{fig:dynamicSeg}
	\end{center}
\end{figure*}
%
\begin{figure*}[t]
	\begin{center}
		\includegraphics[width = 15.9cm]{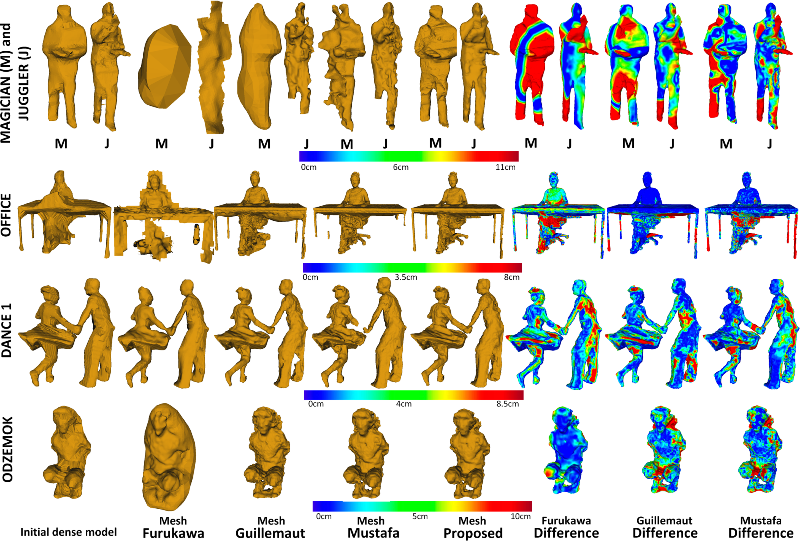}
		\vspace{-0.1cm}
		\caption{Reconstruction result mesh comparison}
		\vspace{-1cm}
		\label{fig:meshes}
	\end{center}
\end{figure*}
\begin{table*}
	\begin{center}
		\begin{tabular}{|c|c|c|c|c|c|c|c|} 
			\hline
			Dataset & Number of Views & Kowdle & Djelouah & Guillemaut & Mustafa & Proposed \\
			\hline
			Couch  & 11 & $99.6 \pm 0.1$ & $99.0 \pm 0.2$ & $97.0 \pm 0.3$ & $98.5 \pm 0.2$ & \textbf{99.7 $\pm$ 0.3}\\
			Chair  & 18 & \textbf{99.2 $\pm$ 0.4} & $98.6 \pm 0.3$ & $97.9 \pm 0.5$  & $98.0 \pm 0.5$ & $99.1 \pm 0.3$ \\
			Car    & 44 & $98.0 \pm 0.7$ & $97.0 \pm 0.8$ & $95.0 \pm 0.7$ & $97.6 \pm 0.3$ & \textbf{98.6 $\pm$ 0.4} \\
			\hline
		\end{tabular}
	\end{center}
	\vspace{-0.3cm}
	\caption{Static segmentation completeness comparison with existing methods on benchmark datasets}
	\vspace{-0.4cm}
	\label{staticSegResults}
\end{table*}
\begin{table}[h]
	\begin{tabular}{|l|l|l|l|l|}
		\hline
		& $\lambda_{data}$ & $\lambda_{c}$ & $\lambda_{smooth}$ & $\lambda_{color}$ \\ \hline
		{\small Magician/Dance2} & 0.4 & 5.0 & .0005 & 0.6 \\ \hline
		Juggler & 0.5 & 5.0 & .0005 & 0.4 \\ \hline
		{\small Odzemok/Dance1/Office} & 0.4 & 3.0 & .001 & 0.6 \\ \hline
	\end{tabular}
	\caption{Parameters used for all datasets: $\lambda_{c}$ represents $\lambda_{contrast}$}
	\label{parameters}
\end{table}

\vspace{-0.6cm}
\section{Results and Performance Evaluation} 
\label{sec:results}
\vspace{-0.1cm}
The proposed system is tested on publicly available multi-view research datasets of indoor and outdoor scenes: static data for segmentation comparison Couch, Chair and Car\cite{Kowdle12}; and dynamic data for full evaluation Dance2\cite{4DInria}, Office\footnotemark, \footnotetext{http://cvssp.org/data/} Dance1\footnotemark[1],  Odzemok\footnotemark[1], Magician and Juggler \cite{UnstructuredVBR10}. Dance1, Dance2 and Office are captured from 8 static cameras, Odzemok from 6 static and 2 moving cameras and  Magician and Juggler from 6 moving handheld cameras. More information is available on the website\footnotemark \footnotetext{http://cvssp.org/projects/4d/4DRecon/}.
%
\begin{figure*}[t]
	\begin{center}
		\includegraphics[width = 15cm]{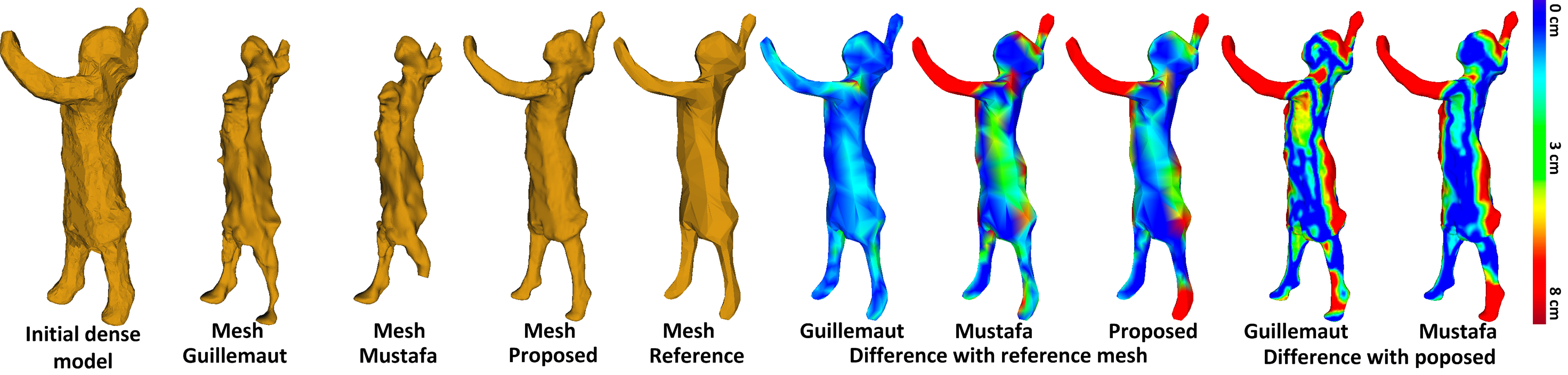}
		\vspace{-0.1cm}
		\caption{Reconstruction result comparison with reference mesh and proposed for Dance2 benchmark dataset}
		\vspace{-0.9cm}
		\label{fig:dance2}
	\end{center}
\end{figure*}
\vspace{-0.2cm}
\subsection{Multi-view segmentation evaluation}
\label{sec:segresults}
\vspace{-0.2cm}
Segmentation is evaluated against the state-of-the-art methods for multi-view segmentation Kowdle\cite{Kowdle12} and Djelouah\cite{Djelouah13} for static scenes and joint segmentation reconstruction per frame Mustafa\cite{MustafaICCV15} and using temporal information Guillemaut\cite{Guillemaut3dv} for both static and dynamic scenes.\\
For static multi-view data the segmentation is initialised as detailed in Section \ref{sec:overview}  followed by refinement using the constrained optimisation Section \ref{sec:optimize}. For dynamic scenes the full pipeline with temporal coherence is used as detailed in \ref{sec:method}.
Ground-truth is obtained by manually labelling the foreground for Office, Dance1 and Odzemok dataset, and for other datasets ground-truth is available online.
We initialize all approaches by the same proposed initial coarse reconstruction for fair comparison.\\
%
%
To evaluate the segmentation we measure completeness as the ratio of intersection to union with ground-truth\cite{Kowdle12}. Comparisons are shown in Table \ref{staticSegResults} and Figure \ref{fig:staticSeg} and \ref{fig:kowdle} for static benchmark datasets and in Table \ref{dynamicSegResults} and Figure \ref{fig:segment} and \ref{fig:dynamicSeg} for dynamic scenes.
Results for multi-view segmentation of static scenes are more accurate than Djelouah, Mustafa and Guillemaut and  comparable to Kowdle with improved segmentation of some detail such as the back of the chair. \\
For dynamic scenes the geodesic star convexity based optimization  together with temporal consistency gives improved segmentation of fine detail such as the legs of the table in the Office dataset and limbs of the person in the Juggler, Magician and Dance2 datasets in Figure \ref{fig:segment} and \ref{fig:dynamicSeg}. This overcomes limitations of previous multi-view per-frame segmentation. 
%
%
%
\begin{table}[t]
	\begin{center}
		\begin{tabular}{|c|c|c|c|c|c|c|c|} 
			\hline
			Dataset & Guillemaut & Mustafa & Proposed \\
			\hline
			Magician & $68.0 \pm 0.7$ & $88.7 \pm 0.5$ & \textbf{91.2 $\pm$ 0.2} \\
			Juggler  & $84.6 \pm 0.6$ & $87.9 \pm 0.6$ & \textbf{93.3 $\pm$ 0.2} \\
			Odzemok  & $90.1 \pm 0.3$ & $89.9 \pm 0.3$ & \textbf{91.8 $\pm$ 0.2} \\
			Dance1   & $99.2 \pm 0.5$ & $99.4 \pm 0.2$ & \textbf{99.5 $\pm$ 0.2} \\
			Office   & $99.3 \pm 0.4$ & $99.0 \pm 0.3$ & \textbf{99.4 $\pm$ 0.2} \\
			Dance2   & $98.6 \pm 0.3$ & $99.0 \pm 0.2$ & \textbf{99.0 $\pm$ 0.2} \\
			\hline
		\end{tabular}
	\end{center}
	\vspace{-0.25cm}
	\caption{Dynamic scene segmentation completeness in \%}
	\vspace{-0.2cm}
	\label{dynamicSegResults}
\end{table}
%
%
\begin{figure}[t]
	\begin{center}
		\includegraphics[width = 7 cm]{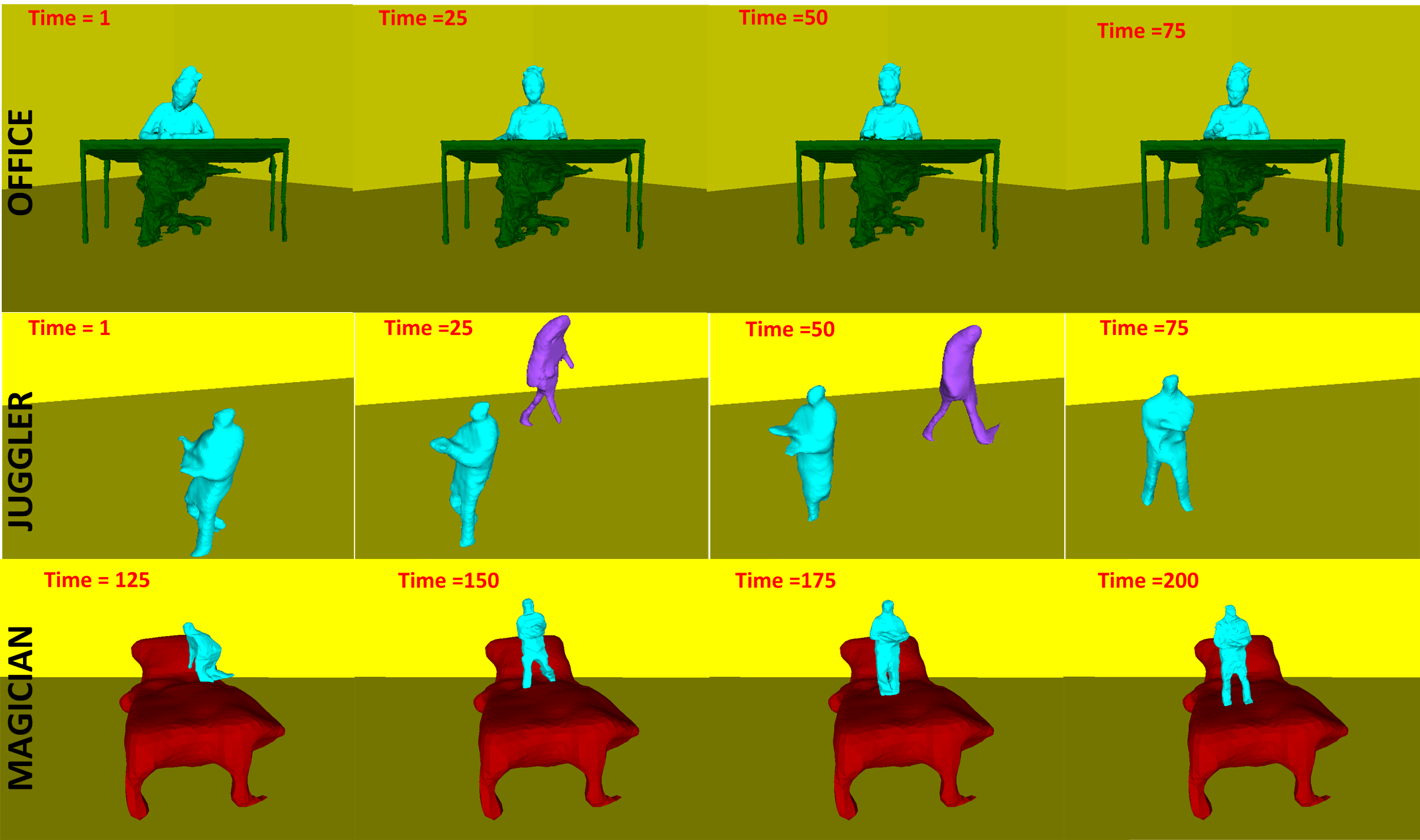}
		\caption{Complete scene reconstruction with 4D mesh sequence.}
		\vspace{-0.5cm}
		\label{fig:CompleteReconst}
	\end{center}
\end{figure}
\vspace{-0.1cm}
\subsection{Reconstruction evaluation}
\vspace{-0.3cm}
Reconstruction results obtained using the proposed method with parameters defined in Table \ref{parameters} are compared against Mustafa\cite{MustafaICCV15}, Guillemaut\cite{Guillemaut3dv}, and Furukawa \cite{Furukawa10} for dynamic sequences. Furukawa \cite{Furukawa10} is a per-frame multi-view wide-baseline stereo approach  which ranks highly on the middlebury benchmark \cite{Seitz06}  but does not refine the segmentation.Figure \ref{fig:meshes} and \ref{fig:dance2} present qualitative and quantitative comparison of our method with the state-of-the-art approaches.
%
%
\begin{table}[h]
	\begin{center}
		\begin{tabular}{|c|c|c|c|c|} 
			\hline
			Dataset & Furukawa & Guillemaut & Mustafa & Ours \\
			\hline
			Dance1   & 326 s & 493 s & 295 s &  \textbf{254 s} \\
			Magician & \textbf{311 s} & 608 s & 377 s &  325 s \\
			Odzemok  & 381 s & 598 s & 394 s &  \textbf{363 s} \\
			Office   & 339 s & 533 s & 347 s &  \textbf{291 s} \\
			Juggler  & 394 s & 634 s & 411 s &  \textbf{378 s} \\
			Dance2   & 312 s & 432 s & 323 s &  \textbf{278 s} \\
			\hline
		\end{tabular}
	\end{center}
	\vspace{-0.25cm}
	\caption{Comparison of computational efficiency for dynamic datasets (time in seconds (s))}
	\vspace{-0.2cm}
	\label{time}
\end{table}
Comparison of reconstructions demonstrates that the proposed method gives consistently more complete and accurate models.
The colour maps highlight the quantitative differences in reconstruction.
As far as we are aware no ground-truth data exist for dynamic scene reconstruction from real multi-view video. 
In Figure \ref{fig:dance2} we present a comparison with the reference mesh available with the Dance2 dataset reconstructed using a visual-hull
approach. This comparison demonstrates improved reconstruction of fine detail with the proposed technique.\\
In contrast to all previous approaches the proposed method gives temporally coherent 4D model reconstructions with dense surface correspondence over time. The introduction of temporal coherence constrains the reconstruction in regions which are ambiguous on a particular frame such as the right leg of the juggler in Figure  \ref{fig:meshes} resulting in more complete shape. Figure  \ref{fig:CompleteReconst} shows three complete scene reconstructions with 4D models of multiple  objects. The Juggler and Magician sequences are reconstructed from moving hand-held cameras. \\
Computation times for the proposed approach vs other methods are presented in Table \ref{time}. 
The proposed approach to reconstruct temporally coherent 4D models is comparable in computation time to per-frame multiple view reconstruction and gives a $\sim$50\% reduction in computation cost compared to previous joint segmentation and reconstruction approaches using a known background. This efficiency is achieved through improved per-frame initialisation based on temporal propagation and the introduction of the geodesic star constraint in joint optimisation. Further results can be found in the supplementary material.
\vspace{-0.3cm}
\section{Conclusion} 
\label{sec:conclusion}
\vspace{-0.2cm}
This paper present a framework for temporally coherent 4D model reconstruction of dynamic scenes from a set of wide-baseline moving cameras. The approach gives a complete model of all static and dynamic non-rigid objects in the scene. Temporal coherence for dynamic objects addresses limitations of previous per-frame reconstruction giving improved reconstruction and segmentation together with dense temporal surface correspondence for dynamic objects.
A sparse-to-dense approach is introduced to establish temporal correspondence for non-rigid objects using robust sparse feature matching to initialise dense optical flow providing an initial segmentation and reconstruction. Joint refinement of object reconstruction and segmentation is then performed using a multiple view optimisation with a novel geodesic star convexity constraint that gives improved shape estimation and is computationally efficient.
Comparison against state-of-the-art techniques for multiple view segmentation and reconstruction demonstrates significant improvement in performance for complex scenes. The approach enables reconstruction of 4D models for complex scenes which has not been demonstrated previously. 
\\
{\bf Limitations:} As with previous dynamic scene reconstruction methods the proposed approach has a number of limitations: persistent ambiguities in appearance between objects will degrade the improvement achieved with temporal coherence; scenes with a large number of inter-occluding dynamic objects will degrade performance; the approach requires sufficient wide-baseline views to cover the scene. 
\noindent \textbf{Acknowledgements:} \\
This research was supported by the European Commission, FP7 Intelligent Management Platform for Advanced Real-time Media Processes project (grant 316564).
\vspace{-0.3cm}
{\small
\bibliographystyle{ieee}
\bibliography{egbib}
\end{document}